\documentclass[runningheads]{llncs}
\usepackage{graphicx}
\usepackage{microtype}
\usepackage{enumerate}
\usepackage{graphicx}
\usepackage{subfigure}
\usepackage{booktabs} 
\usepackage{mathrsfs}  
\usepackage{amssymb,amsmath}
\usepackage{caption}
\usepackage{wrapfig}
\usepackage{picinpar}
\usepackage{cutwin}
\usepackage{algorithm}  
\usepackage{algorithmic}
\usepackage{amsfonts,amssymb}
\usepackage{verbatim}
\usepackage{sidecap}
\usepackage{color}
\usepackage{url}
\DeclareMathOperator{\Exp}{Exp}
\DeclareMathOperator{\Log}{Log}
\DeclareMathOperator{\Dist}{Dist}

\usepackage{appendix}
\usepackage{hyperref}

\begin{document}
\title{Mixture Probabilistic Principal Geodesic Analysis}
\titlerunning{MPPGA}
%

\author{Youshan Zhang\inst{1} \and Jiarui Xing\inst{2} \and Miaomiao Zhang\inst{2,3}}
%

\authorrunning{Y. Zhang et al.}
%

\institute{Computer Science \& Engineering, Lehigh University, Bethlehem, USA 
\and Electrical \& Computer Engineering, University of Virginia, Charlottesville, USA
\and Computer Science, University of Virginia, Charlottesville, USA}

%
\maketitle              

\begin{abstract}
Dimensionality reduction on Riemannian manifolds is challenging due to the complex nonlinear data structures. While probabilistic principal geodesic analysis~(PPGA) has been proposed to generalize conventional principal component analysis (PCA) onto manifolds, its effectiveness is limited to data with a single modality. In this paper, we present a novel Gaussian latent variable model that provides a unique way to integrate multiple PGA models into a maximum-likelihood framework. This leads to a well-defined mixture model of probabilistic principal geodesic analysis (MPPGA) on sub-populations, where parameters of the principal subspaces are automatically estimated by employing an Expectation Maximization algorithm. We further develop a mixture Bayesian PGA (MBPGA) model that automatically reduces data dimensionality by suppressing irrelevant principal geodesics. We demonstrate the advantages of our model in the contexts of clustering and statistical shape analysis, using synthetic sphere data, real corpus callosum, and mandible data from human brain magnetic resonance~(MR) and CT images.

\end{abstract}
%
%
%
\section{Introduction}
PCA has been widely used to analyze high-dimensional data due to its effectiveness in finding the most important principal modes for data representation ~\cite{jolliffe1986principal}. Motivated by the nice properties of probabilistic modeling, a latent variable model of PCA for factor analysis was presented~\cite{tipping1999probabilistic,roweis1998algorithms}. Later, different 
variants of probabilistic PCA including Bayesian PCA~\cite{bishop1999bayesian} and
mixture models of PCA~\cite{chen1999mixture} were developed for automatic
data dimensionality reduction and clustering, respectively. It is important to extend all
these models from flat Euclidean spaces to general Riemannian manifolds, where
the data is typically equipped with smooth constraints. For instance, an appropriate
representation of directional data, i.e., vectors of unit length
in $R^n$, is the sphere $S^{n-1}$~\cite{mardia2009directional}. Another important example
of manifold data is in shape analysis, where the definition of the shape of an object
should not depend on its position, orientation, or scale, i.e., Kendall shape space~\cite{kendall1984shape}.
Other examples of manifold data include geometric transformations such as rotations
and translations, symmetric positive-definite tensors~\cite{fletcher2004principal,tuzel2008pedestrian}, Grassmannian manifolds (a set of $m$-dimensional linear subspaces of $R^n$),
and Stiefel manifolds (the set of orthonormal $m$-frames in $R^n$)~\cite{turaga2011statistical}.

Data dimensionality reduction on manifolds is challenging due to the
commonly used linear operations violate the natural constraints of manifold-valued data.
In addition, basic statistical terms such as distance metrics, or data distributions
vary on different types of manifolds~\cite{kendall1984shape,turaga2011statistical,obata1962certain}.
A groundbreaking work, known as principal geodesic analysis (PGA), was the first to
generalize PCA to nonlinear manifolds~\cite{fletcher2004principal}. This method describes the geometric variability of manifold data by finding lower-dimensional geodesic subspaces that minimize the residual sum-of-squared geodesic distances to the data. Later on, an exact solution to
PGA~\cite{sommer2010manifold,sommer2014optimization} and a robust formulation for estimating the output results~\cite{banerjee2017robust} were developed.
The probabilistic interpretation of PGA was firstly introduced in~\cite{zhang2013probabilistic}, which paved a way for factor analysis on manifolds. Since PPGA only defines a single projection of the data, the scope of its application is limited to uni-modal distributions. A more natural and motivating solution is to model the multi-modal data structure with a collection or mixture of local sub-models. Current mixture models on a specific manifold generally employ a two-stage procedure: a clustering of the data projected in Euclidean space followed by performing PCA within each cluster~\cite{cootes1999mixture}. None of these algorithms define a probability density.

In this paper, we derive a mixture of PGA models
as a natural extension of PPGA~\cite{zhang2013probabilistic}, where all model parameters including the low-dimensional factors
for each data cluster is estimated through the maximization of a single likelihood function. The theoretical foundation of developing generative models of principal geodesic analysis for multi-population studies on general manifolds is brand new. In addition, the algorithmic inference of our proposed method is nontrivial due to the complicated geometry of manifold-valued data and numerical issues. Compared to previous methods,
the major advantages of our model are: (i) it leads to a unified algorithm that well integrates soft data clustering and principal subspaces estimation on general Riemannian manifolds; (ii) in contrast to the two-stage approach mentioned above, our model explicitly considers the reconstruction error of principal modes as a criterion for clustering tasks; and (iii) it provides a more powerful way to learn features from data in non-Euclidean spaces with multiple subpopulations. We showcase our model advantages from two distinct perspectives: automatic data clustering and dimensionality reduction for analyzing shape variability. In order to validate the effectiveness of the proposed algorithm, we compare its performance with the state-of-the-art methods on both synthetic and real datasets. We also briefly discuss a Bayesian version of our mixture PPGA model that equips with the functionality of automatic dimensionality selection on general manifold data.



\section{Background: Riemannian Geometry and PPGA}
In this section, we briefly review PPGA~\cite{zhang2013probabilistic} defined on a smooth Riemannian manifold $M$, which is a
generalization of PPCA~\cite{tipping1999probabilistic} in Euclidean space. Before introducing the
model, we first recap a few basic concepts of Riemannian geometry (more details are provided in~\cite{do1992riemannian}). 

\textbf{Covariant Derivative.} The covariant derivative is a generalization of
the Euclidean directional derivative to the manifold setting. 
Consider a curve $c(t): [0,1] \rightarrow M$ and let $\dot{c} = dc/dt$ be its
velocity. Given a vector field $V(t)$ defined along $c$, we can define the
covariant derivative of $V$ to be $\frac{DV}{dt} = \nabla_{\dot{c}} V$ that 
reflects the change of the vector field $\dot{c}$ in the $V$ direction. 
A vector field is called parallel if the covariant derivative along the curve
$c$ is zero. A curve $c$ is geodesic if it satisfies the equation
$\nabla_{\dot{c}} \dot{c} = 0$.

\textbf{Exponential Map.} For any point $p \in M$ and tangent vector $v \in T_p M$ (also known as the
tangent space of $M$ at $p$), there exists a unique geodesic curve $c$ with
initial conditions $c(0) = p$ and $\dot{c}(0) = v$. This geodesic is
only guaranteed to exist locally. The Riemannian exponential map at $p$ is defined as $\Exp_p(v) = c(1)$. In other words, the exponential map takes a position and velocity as
input and returns the point at time $t=1$ along the geodesic with certain initial
conditions. Notice that the exponential map is simply an addition in Euclidean space, 
i.e., $\Exp_p(v) = p + v$. 

\textbf{Logarithmic Map.} The exponential map is locally diffeomorphic onto a neighborhood of
$p$. Let $V(p)$ be the largest such neighborhood, the Riemannian log map, $\Log_p: V(p) \rightarrow T_p M$, is an inverse of the exponential map within $V(p)$. For any point $q \in V(p)$, the
Riemannian distance function is given by $\Dist(p, q) = \| \Log_p(q)\|$. Similar to the exponential 
map, this logarithmic map is a subtraction in Euclidean space, i.e., $\Log_p(q) = q - p$.   
\subsection{PPGA}
Given an $d$-dimensional random variable $y \in M$, the main idea of PPGA~\cite{zhang2013probabilistic}
is to model $y$ as 
\begin{equation}\label{eq:ppga}
y = \Exp(\, \Exp(\mu, Bx), \epsilon \, ), \quad B = W \Lambda,
\end{equation} 
where $\mu$ is a base point on $M$, $x \in \mathbb{R}^q$ is a $q$-dimensional latent variable, with
$x \sim N(0, I)$, $B$ is an $d \times q$ factor matrix that relates $x$ and $y$,
and $\epsilon$ represents error. We will find it is convenient
to model the factors as $B = W \Lambda$, where $W$ is a matrix with $q$ columns
of mutually orthogonal tangent vectors in $T_\mu M$, $\Lambda$ is a $q \times q$ diagonal
matrix of scale factors for the columns of $W$. This removes
the rotation ambiguity of the latent factors and makes them analagous to the
eigenvectors and eigenvalues of standard PCA (there is still of course an
ambiguity of the ordering of the factors). 

The likelihood of PPGA is defined by a generalization of the normal distribution $\mathcal{N}(\mu, \tau^{-1})$, called Riemannian normal distribution, with its precision parameter $\tau$. 
Therefore, we have
\begin{align}\label{eq:Nr}
p(y | \mu, \tau) &= \frac{1}{C(\mu, \tau)} \exp\left( - \frac{\tau}{2} \Dist(y, \mu)^2 \right), \quad \text{with} \nonumber \\
C(\mu, \tau) &= \int_M \exp\left( -\frac{\tau}{2} \Dist(y, \mu)^2 \right) dy.
\end{align}

This distribution is applicable to any Riemannian manifold, and the value of $C$ in Eq.~\ref{eq:Nr} does not depend on $\mu$. It 
reduces to a multivariate normal distribution with isotropic
covariance when $M = \mathbb{R}^n$ (see~\cite{fletcher2013geodesic} for details). 
Note that this noise model could be replaced with other different distributions
according to different types of applications.

Now, the PPGA model for a random variable $y$ in Eq.~\eqref{eq:ppga} can be defined as
\begin{align}\label{eq:model}
y \sim \mathcal{N} \left( \Exp(\mu, s), \tau^{-1} \right), \, s = W \Lambda x. 
\end{align}

\section{Our Model: Mixture Probability Principal Geodesic Analysis (MPPGA)}
We now introduce a mixture model of PPGA (MPPGA) that provides a tempting prospect of being able to model complex multi-modal data structures. This formulation allows all model parameters to be estimated from maximum-likelihood, where both an appropriate data clustering and the associated principal modes are jointly optimized. 

\begin{figure}[h]
\centering
\includegraphics[scale=0.5]{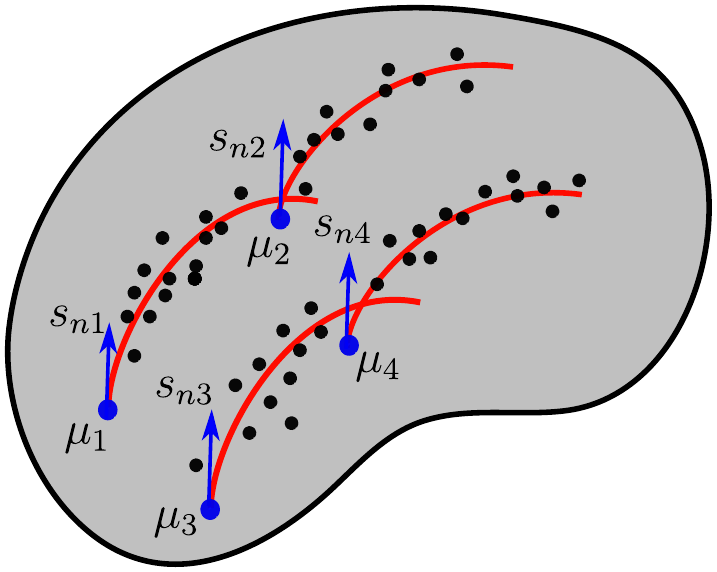}\\
\caption{Example MPPGA model with four clusters.}
\label{fig:four}
\end{figure}

Consider observed data $y_n \in \{y_1, \cdots, y_N \}$ generated from $K$ clusters
on $M$ (as shown in Fig.~\ref{fig:four}). We first introduce a $K$-dimensional binary random variable $z_n$ with its $k$-th element $z_{nk} \in \{0, 1\}$ as an indicator for $n$-th data point that belongs to cluster $k$, where $k \in \{1, \cdots, K\}$. This indicates that $z_{nk}=1$ with other value being zero if the data $y_{n}$ is in cluster $k$. The probability of each random variable $z_n$ is
\begin{equation}\label{eq:pz}
  p(z_n)=\prod_{k=1}^{K}\pi_{k}^{z_{nk}},    
\end{equation}
where $\pi_{k} \in [0, 1]$ is the model mixing coefficient that satisfies $\sum\limits _{k=1}^{K}\pi_{k}=1$. 

Analogous to PPGA in Eq.~\eqref{eq:ppga}, the likelihood of each observed data $y_n$ is
\begin{align}\label{eq:p|z}
  p(y_n \, | \, z_n) &=\prod_{k=1}^{K} \mathcal{N}(y_n \, | \, \Exp(\mu_k,s_{nk}),\tau_{k}^{-1})^{z_{nk}}, \quad \text{with} \nonumber \\ 
  s_{nk} &=W_{k} \Lambda_{k} x_{nk},
\end{align}
where $x_{nk} \sim  \mathcal{N}(0,I)$ is a latent random variable in $\mathbb{R}^{q}$, $\mu_k$ is a
base point for each cluster $k$, $W_k$ is a matrix with each columns representing the mutually orthogonal
tangent vectors in $T_{\mu_k}M$, and $\Lambda_k$ is a diagonal matrix of
scale factors for the columns of $W_k$.

Combining Eq.~\eqref{eq:pz} with Eq.~\eqref{eq:p|z}, we obtain the complete data likelihood
\begin{align}\label{eq:p,z}
\centering
  p(y,z) & = \prod_{n=1}^{N} p(y_n \, | \, z_n)p(z_n) p(x_n)\nonumber \\
   & =\prod_{n, k=1}^{N, K} [\pi_{k} p(y_n \, | \, \Exp(\mu_k,s_{nk}),\tau_{k}^{-1}) p(x_{nk})]^{z_{nk}}.
\end{align}

The log of the data likelihood in Eq.~\eqref{eq:p,z} can be computed as
 \begin{align}\label{eq:L}
\centering
 \mathcal{L} \triangleq \ln \,  p(y,z) = -\sum_{n, k=1}^{N, K} z_{nk} \,  \ln  \lbrace \pi_{k} p(y_n \,|\, \Exp(\mu_k,s_{nk}),\tau_{k}^{-1}) p(x_{nk}) \rbrace.
\end{align}

\subsection{Inference}\label{sec:inference}
We employ a maximum likelihood expectation maximization (EM) method to estimate model parameters $\theta=(\pi_k, \mu_k, W_k, \Lambda_k, \tau_k, x_{nk})$ and latent variables $z_{nk}$. This scheme includes two main steps:
\paragraph*{E-step.} To treat the binary indicator $z_{nk}$ fully as latent
random variables, we integrate them out from the distribution defined in Eq.~\eqref{eq:p,z}. 
Similar to typical Gaussian mixture models, the expectation value of
the complete-data log likelihood function is 
\begin{align} \label{eq:EL}
   \mathbb{E} [\mathcal{L}] =  - \sum_{n, k=1}^{N, K}\mathbb{E}[z_{nk}] \, \{\ln \, p(y_n \,|\, \Exp(\mu_k,s_{nk}),\tau_{k}^{-1})
    + \ln\, p(x_{nk}) + \ln \pi_{k} \}.
\end{align}

The expected value of the latent variable $z_{nk}$, also known as the responsibility
of component $k$ for data point $y_n$~\cite{bishop2006pattern}, is then computed by its posterior 
distribution as
\begin{align}\label{eq:gamma}
\centering
\mathbb{E}[z_{nk}] = p(z_{nk} | y_{n})&= \frac{p(y_{n}|z_{nk})p(z_{nk})}{\sum_{k=1}^{K}p(y_{n}|z_{nk})p(z_{nk})} \nonumber \\
&=\frac{\pi_{k}p (y_{n}|\Exp(\mu_k,z_{nk}),\tau_{k}^{-1})}{\sum_{k=1}^{K}\pi_{k}p(y_{n}|\Exp(\mu_k,z_{nk}),\tau_{k}^{-1})}.
\end{align}

Recall that the Rimannian distance function $\Dist(p, q) = \| \Log_p(q)\|$.
We let $\gamma_{nk} \triangleq \mathbb{E}[z_{nk}]$ and rewrite Eq.~\eqref{eq:EL} as
\begin{align}\label{eq:ELformula}
 \mathbb{E} [\mathcal{L}] = -\sum_{n, k=1}^{N, K} \gamma_{nk} 
 \{ \frac{\tau_{k}}{2} \, \Log(\Exp(\mu_k,s_{nk}),y_{n})^2 
  + \ln C + \ln \pi_{k} +\frac{||x_{nk}||^2}{2} \},
\end{align}
where $C$ is a normalizing constant.  
 
\paragraph*{M-step.} We use gradient ascent to maximize the expectation function 
$\mathbb{E} [\mathcal{L}]$ and update parameters $\theta$. 
Since the maximization of the mixing coefficient $\pi_k$ is the
same as Gaussian mixture model~\cite{bishop2006pattern}, we only 
give its final close-form update here as $\tilde{\pi}_{k}=\sum_{n=1}^{N}  \gamma_{nk} / N$.

The computation of the gradient term requires we compute the 
derivative operator (Jacobian matrix) of the exponential map, i.e.,
$d_{\mu_k} \Exp(\mu_k, s_{nk})$, or $d_{s_{nk}} \Exp(\mu_k, s_{nk})$.
Next, we briefly review the computations of derivatives w.r.t. the mean point $\mu$ and the tangent vector $s$ separately. Closed-form formulations of these derivatives in the space of sphere, or 2D Kendall shape space are provided in~\cite{zhang2013probabilistic,fletcher2016probabilistic}. 

\begin{figure}[!h]
\centering
\includegraphics[scale=1.0]{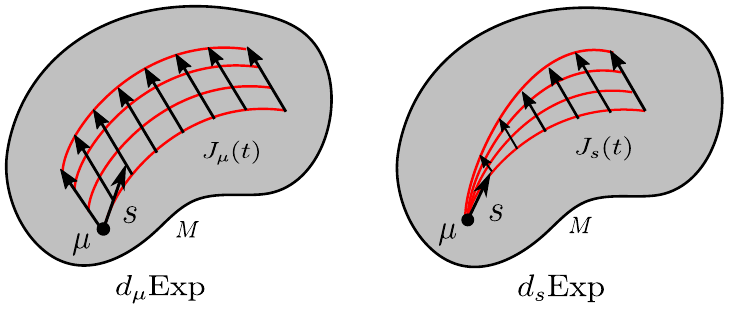}
\caption{ Jacobi fields}
\label{fig:Ja}
\end{figure}

\sloppy
\paragraph*{For derivative w.r.t. $\mu$.} Consider a variation of geodesics, e.g., $c(h, t) = \Exp(\Exp(\mu, hu), ts(h))$, where $u \in T_{\mu}M$ and $s(h)$ comes from parallel translating $s$ along the geodesic $\Exp(\mu, hu)$. The derivative of this variation results in a Jacobi field: $J_{\mu}(t) = dc/dh(0, t)$. This gives an expression for the exponential map derivative as $d_{\mu} \Exp(\mu, s) = J_{\mu}(1)$ (as shown on the left panel of Fig.~\ref{fig:Ja}).

\paragraph*{For derivative w.r.t. $s$.} Consider a variation of geodesics, e.g., $c(h, t) = \Exp(\mu, hu + ts)$. Again, the derivative of the exponential map is given by a Jacobi field satisfying $J_s(t) = dc/dh(0, t)$, and we have $d_s \Exp(\mu, s) u = J_s(1)$ (as shown on the right panel of Fig.~\ref{fig:Ja}).

Now we are ready to derive all gradient terms of $\mathbb{E} [\mathcal{L}]$ in 
Eq.~\ref{eq:ELformula} w.r.t. the parameters $\theta$. For purpose of better readability, we simplify the notation by defining
$\Log(\cdot) \triangleq \Log\left(\Exp(\mu_{k}, s_{nk}), y_n\right)$
in remaining sections.

\paragraph*{Gradient for $\mu_k$:} the gradient of updating $\mu_k$ is
\begin{equation}\label{eq:gmu}
\nabla_{\mu_k}\mathbb{E} [\mathcal{L}]=\sum_{n, k=1}^{N, K}  
\gamma_{nk} \, \tau_k \, d_{\mu_k} \Exp(\mu_k, s_{nk})^{\dagger} \Log(\cdot),
\end{equation}
where $\dagger$ represents adjoint operator, i.e., for any tangent vectors $\hat{u}$
and $\hat{v}$,
$$\langle d_{\mu_k} \Exp(\mu_k, s_{nk})\hat{u}, \hat{v} \rangle = \langle \hat{u}, d_{\mu_k} \Exp(\mu_k, s_{nk})^{\dagger} \hat{v}\rangle.$$

\paragraph*{Gradient for $\tau_k$:} the gradient of $\tau_k$ is computed as
\begin{align}\label{eq:gtau}
\nabla_{\tau_k} \mathbb{E} [\mathcal{L}] &=\sum_{n, k=1}^{N, K}   \gamma_{nk}\frac{1}{C(\tau)}A_{n-1}\int_{0}^{R}\frac{r^2}{2}\text{Exp}(-\frac{\tau}{2}r^2) \cdot \nonumber \\ 
&\prod_{\kappa=2}^{n}\kappa_{\kappa}^{-1/2}f_{\kappa}(\sqrt{\kappa_{\kappa}}r)dr-\frac{1}{2}\Log(\cdot)^{2} dr,
\end{align}
where $A_{n-1}$ is the surface area of $n-1$ hypershpere. $r$ is radius, $\kappa_\kappa$ is the sectional curvature. Here $R=\text{min}_{v}{R(v)} $, where $R(v)$ is the maximum distance of $\text{Exp}(\mu_k, rv)$ with $v$ being a point of unit sphere $S^{n-1} \subset T_{\mu_k} M$. While this formula is only valid for simple connected symmetric spaces, other spaces should be changed according to different definitions of the probability density function in Eq.~\eqref{eq:Nr}.

To derive the gradient w.r.t. $W_k, \Lambda_k$ and $x_{nk}$, we need to compute 
$d (\Log(\cdot)^2) / d s_{nk}$ first. Analogous to Eq.~\ref{eq:gmu}, we have
\begin{equation}
\frac{d (\Log(\cdot)^2)}{d s_{nk}} = 2 \left(d_{s_{nk}} \Exp(\mu_k, s_{nk})^\dag \Log(\cdot) \right).
\end{equation} 

After applying chain rule, we finally get all gradient terms as following:
\paragraph*{Gradient for $W_k$:} the gradient term of $W_k$ is
\begin{equation}\label{eq:gw}
\nabla_{W_k} \mathbb{E} [\mathcal{L}] = \sum_{n, k=1}^{N, K}   \gamma_{nk}\, 
 \frac{\tau_k}{2} \cdot \frac{d (\Log(\cdot)^2)}{d s_{nk}} \cdot \ x_{nk}^{T}\Lambda_k.
\end{equation}
To maintain the mutual orthogonality of each column of $W_k$, we consider $W_k$ as a point in Stiefel manifold $V_q(T_{\mu}M)$, i.e., the space of orthonormal $q$-frames in $T_{\mu}M$, and project the gradient of Eq. \ref{eq:gw} into tangent space $T_{W_k} V_q(T_{\mu}M)$. We then update $W_k$ by taking a small step along the geodesic in the projected gradient direction.  For details on Stiefel manifold, see \cite{edelman1998geometry}.

\paragraph*{Gradient for $\Lambda_k^a$:} the gradient term of each $a$-th diagonal element of $\Lambda_k$ is:
\begin{equation}\label{eq:glambda}
\nabla_{\Lambda_k^a} \mathbb{E} [\mathcal{L}] = \sum_{n, k=1}^{N, K} \gamma_{nk}\, \tau_k(W_k^ax_{nk}^{a})^T \cdot \frac{d (\Log(\cdot)^2)}{d s_{nk}},
\end{equation}
where  $W_k^{a}$ is the $a$th column of $W_k$ and $x_{nk}^{a}$ is the $a$th component of $x_{nk}$ .

\paragraph*{Gradient for $x_{nk}$:} the gradient w.r.t. each $x_{nk}$ is
\begin{equation}\label{eq:gx}
\nabla_{x_{nk}} \mathbb{E} [\mathcal{L}] =
-\sum_{n, k=1}^{N, K} \gamma_{nk} \{ x_{nk}-\frac{\tau_k}{2} \Lambda_k  W_k^{T} \cdot
\frac{d (\Log(\cdot)^2)}{d s_{nk}} \}.
\end{equation}


In this section, we further develop a Bayesian variant of MPPGA that equips 
with the functionality of automatic data dimensionality reduction. 
A critical issue in maximum likelihood estimate of principal geodesic analysis
is the choice of the number of principal geodesic to be retained. 
This also could be problematic in our proposed MPPGA model since
we assume each cluster has different dimensions of principal subspaces, and
an exhaustive search over the parameter space can become computationally intractable. 

To address this issue, we develop a Bayesian mixture principal geodesic 
analysis (MBPGA) model that determines the number of principal modes
automatically to avoid adhoc parameter tuning. We carefully introduces
an automatic relevance determination (ARD) prior~\cite{bishop2006pattern}
on each $a$th diagonal element of the eigenvalue matrix $\Lambda$ as 
\begin{equation}
\label{eq:lambda,alpha}
p(\Lambda | \beta)=\prod_{i=1}^{d-1} (\frac{\beta^a}{2\pi})^{d/2}  e^{-\frac{1}{2}\beta^a \| \Lambda^a\|^2}.
\end{equation}
Each hyper-parameter $\beta^a$ controls the inverse variance
of its corresponding principal geodesic $W^a$, which is the $a$th column of 
$W$ matrix. This indicates that if $\beta^a$ is particularly large, the 
corresponding $W^a$ will tend to be small and will be effectively eliminated.

Incorporating this ARD prior into our MPPGA model defined in Eq.~\ref{eq:L}, 
we arrive at a log posterior distribution of $\Lambda$ as
\begin{equation}\label{eq:lambda,Y}
\ln{p(\Lambda|Y)}= \mathcal{L}-\frac{1}{2}\sum_{i=1}^{d-1}\beta^a \| \Lambda^a \|^{2} + \text{const.}.
\end{equation}

Analogous to the EM algorithm introduced in Sec.~\ref{sec:inference}, we maximize
over $\Lambda^a$ in M-step by using the following gradient:
\begin{equation}\label{eq:B_glambda}
\nabla_{\Lambda^a} \mathbb{E} [\mathcal{L}] = \sum_{n, k=1}^{N, K} \gamma_{nk}\, \tau_k(W_k^ax_{nk}^{a})^T \cdot \frac{d (\Log(\cdot)^2)}{d s_{nk}}-\beta^{a}\Lambda^{a}.
\end{equation}
Similar to the ARD prior discussed in~\cite{bishop1999bayesian}, the hyper-parameter
$\beta^a$ can be effectively estimated by $\beta^a=d / \|\Lambda^a\|^2$, where $d$ is the 
dimension of the original data space. 
\section{Evaluation}
We demonstrate the effectiveness of our MPPGA and MBPGA model by using both synthetic data and real data, and compare with two baseline methods K-means-PCA~\cite{cootes1999mixture} and MPPCA~\cite{tipping1999mixtures} designed for multimodal Euclidean data. The geometry background of specific sphere and Kendall shape space including the computations of Riemannian exponential map, log map, and Jacobi fields can be found in~\cite{zhang2013probabilistic,fletcher2013geodesic}.

\subsection{Data}
\paragraph*{Sphere.} Using the generative model for PGA, we simulate a random sample of $764$ data points on the unit sphere $S^{2} $ with known parameters 
$W, \Lambda, \tau$, and $\pi$ (see Tab~\ref{tab:my_label}). All data points consist
three clusters (Green: $200$; Blue: $289$; Black: $275$). Note that our
ground truth $\mu$ is generated from random uniform points on the sphere. 
The $W$ is generated from a random Gaussian matrix, to which we then apply the Gram-Schmidt
algorithm to ensure its columns are orthonormal.

\paragraph*{Corpus callosum shape.} The corpus callosum data are derived from public released Open Access Series of Imaging Studies (OASIS) database~\url{www.oasis-brains.org}. It includes $32$ magnetic resonance imaging scans of human brain subjects, with age from $19$ to $90$.
The corpus callosum is segmented in a midsagittal slice using the ITK SNAP program~\url{www.itksnap.org}. The boundaries of these segmentations are sampled with $64$ points. This algorithm generates a sampling
of a set of shape boundaries while enforcing correspondences between different point models within the population.

\paragraph*{Mandible shape.} The mandible data is extracted from a collection of CT scans of human mandibles, with $77$ subjects ($36$ female vs. $41$ male) aged from $0$ to $19$. We sample $2 \times 400$ points on the boundaries.




\subsection{Experiments}
We first run our EM algorithm estimation of both MPPGA
and MBPGA to test whether we could recover the model parameters. 
To initialize the model parameters (e.g., the cluster mean $\mu$, principal 
eigenvector matrix $W$, and eigenvalue $\Lambda$), we use the output of K-means algorithm followed 
by performing linear PCA within each cluster. We uniformly
distribute the weight to each mixing coefficient, i.e., $\pi_k = 1/K$. The
initialization of all precision parameters $\{\tau_k\}$ is $0.01$. 
We compare our model with two existing algorithms - mixture probabilistic principal components~(MPPCA)~\cite{tipping1999mixtures} and 
K-means-PCA~\cite{cootes1999mixture} performed in Euclidean space. 
For fair comparison, we keep the number of 
clusters the same across all algorithms.

To further investigate the applicability of our model MPPGA to real data, we
test on 2D shapes of corpus callosum to study brain degeneration. 
The idea is to identify shape differences between two
sub-populations: healthy vs. control group by analyzing 
their shape variability. We also run the extended
Bayesian version of our model MBPGA to automatically select 
a compact set of principal geodesics to represent data variability.
We perform similar experiments on the 2D mandible shape data to 
study group differences across genders, as well as within-group 
shape variability that reflects localized regions of growth.

\subsection{Results}
\begin{figure}[h]
\centering
\subfigure[K-means-PCA]{\label{fig:sphere.a}
\includegraphics[width=0.2\columnwidth]{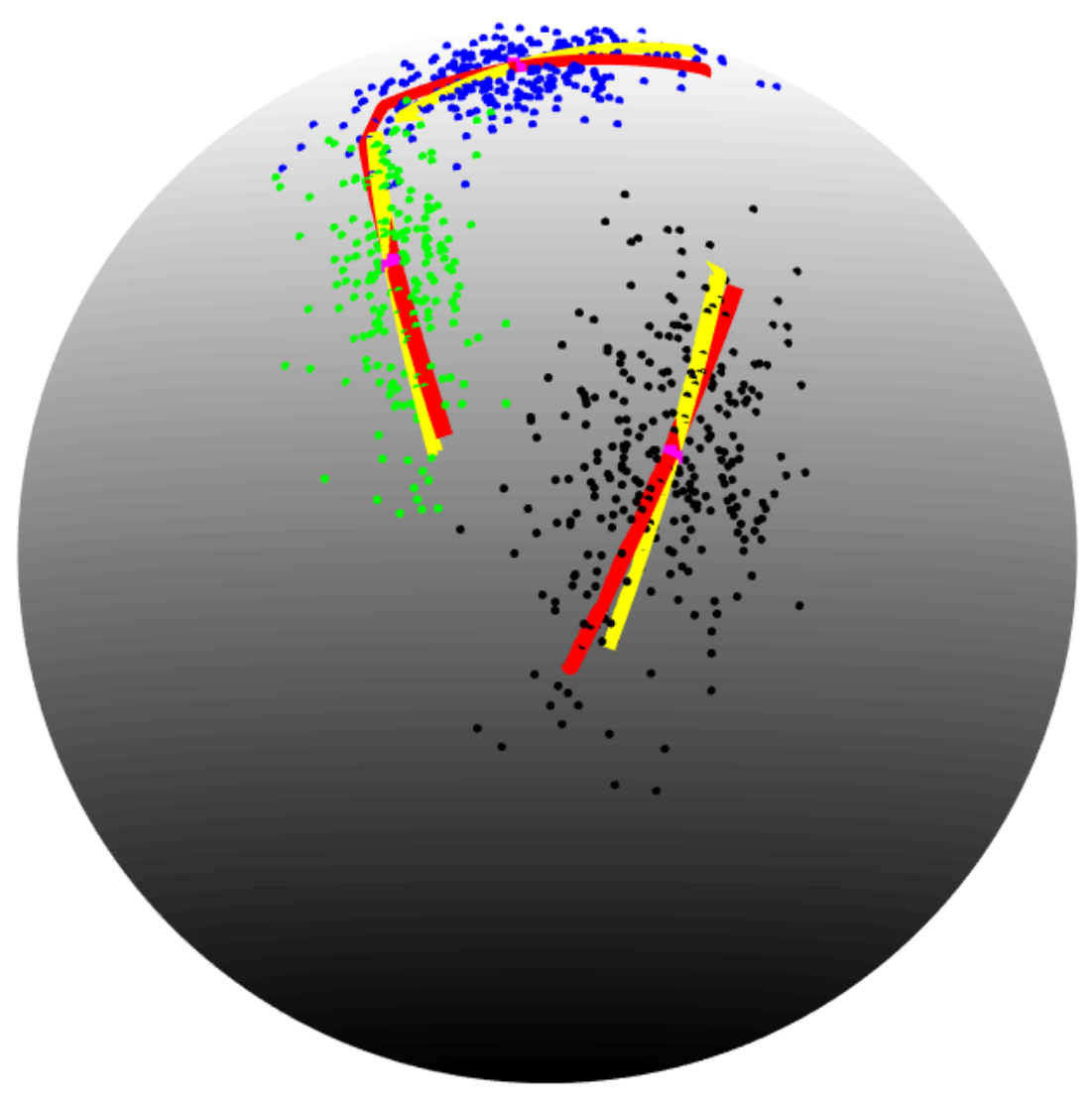}}
\centering
\subfigure[MPPCA]{\label{fig:sphere.b}
\includegraphics[width=0.2\columnwidth]{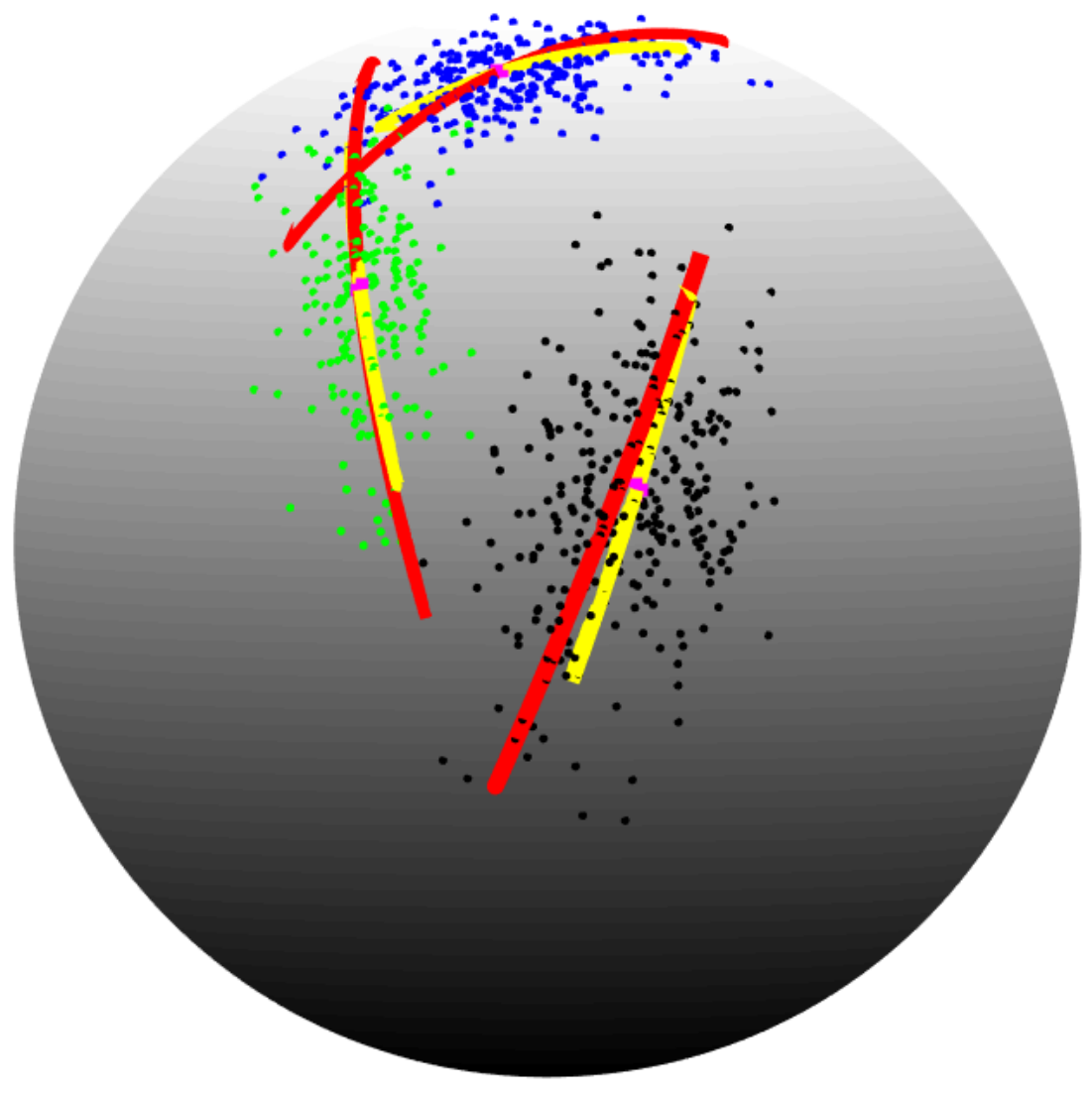}}
\centering
\subfigure[MPPGA]{\label{fig:sphere_c}
\includegraphics[width=0.2\columnwidth]{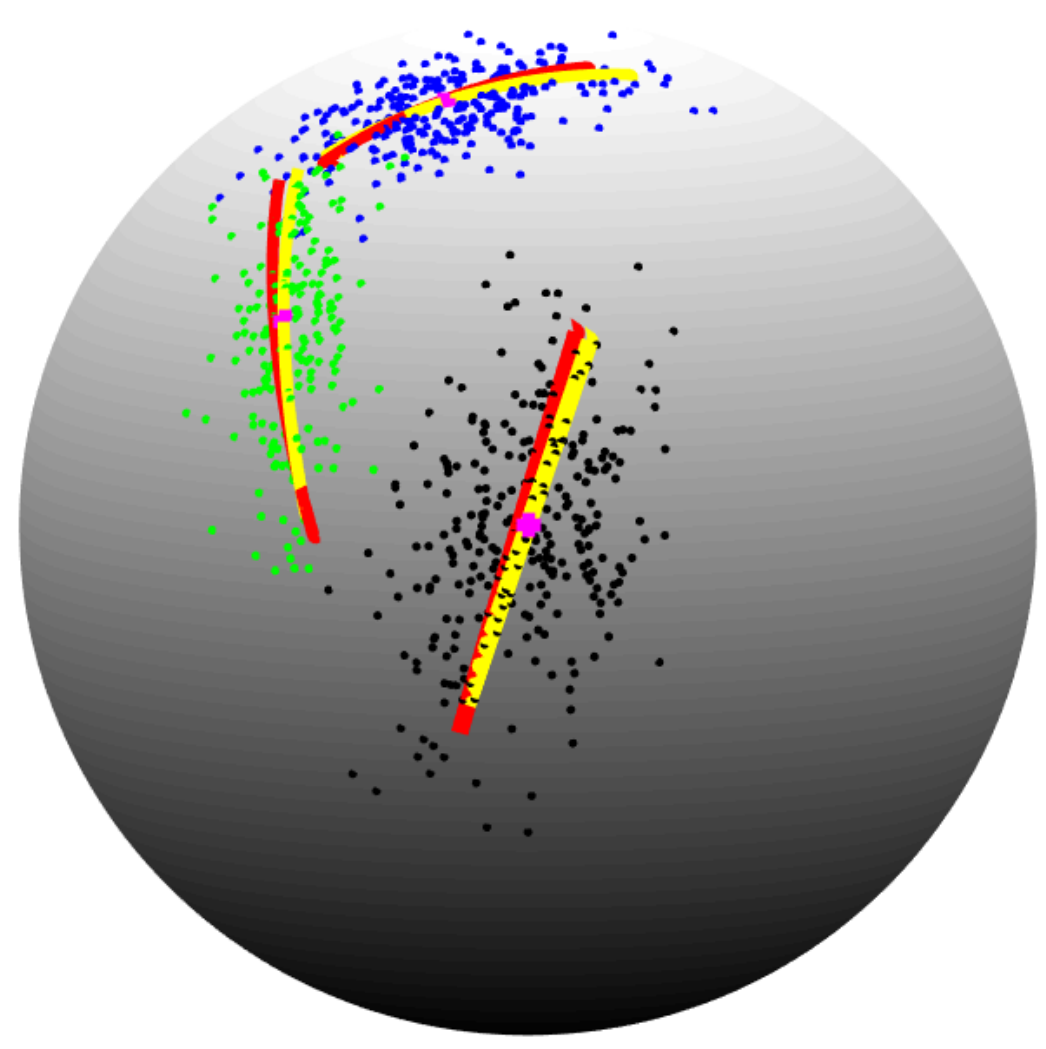}}
\centering
\subfigure[MBPGA]{\label{fig:sphere_d}
\includegraphics[width=0.2\columnwidth]{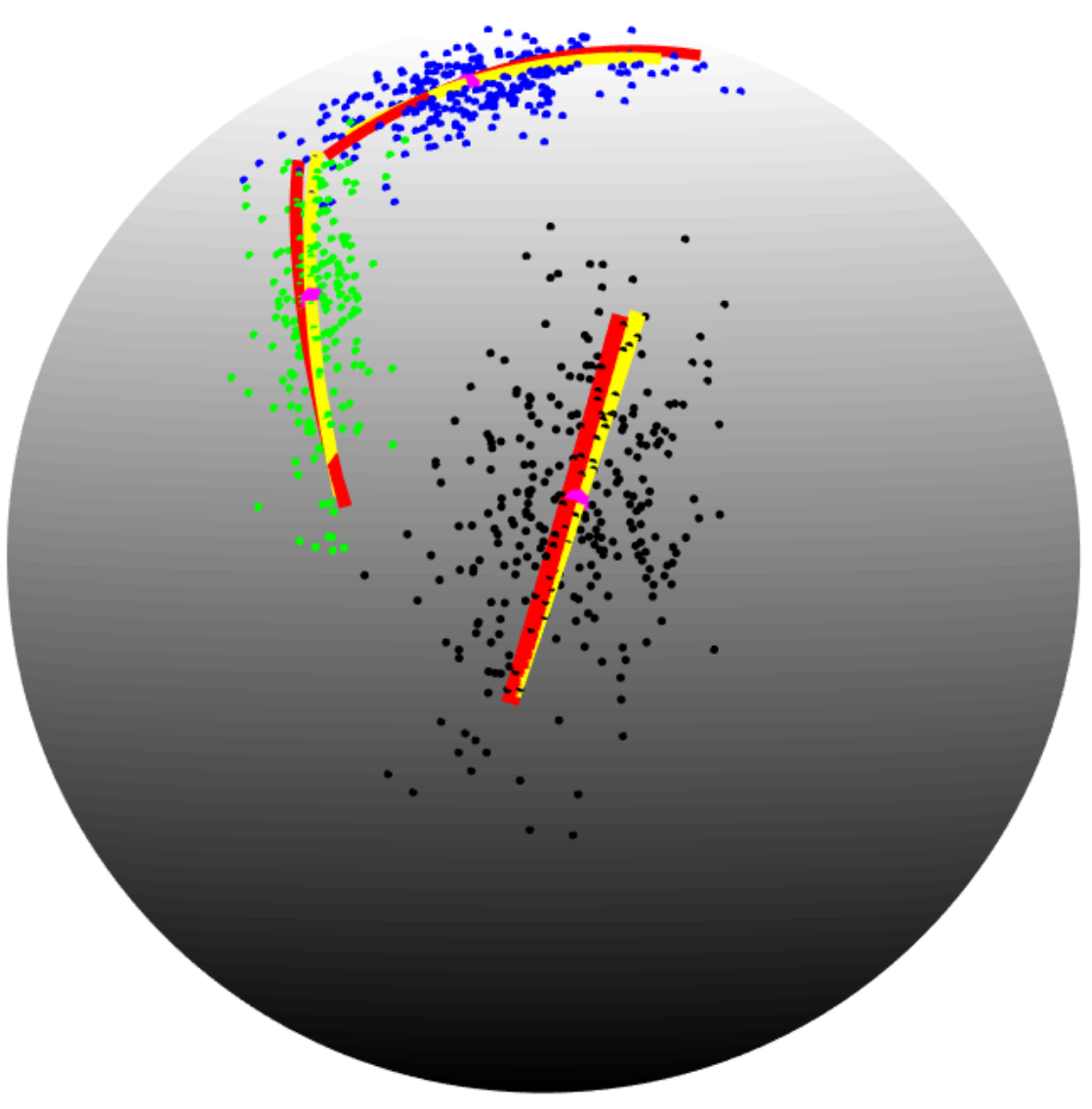}}
\caption{The comparison of our model MPPGA/MBPGA with K-means-PCA and MPPCA (after being projected from Eucliean space onto the sphere). 
We have three clusters marked in green, blue, and black. Yellow lines: ground
truth geodesics; Red lines: estimated geodesics. }
\label{fig:sphere}
\end{figure}

Fig.~\ref{fig:sphere} compares the estimated results of our model MPPGA/MBPGA with two baseline methods K-means-PCA and MPPCA. For the purpose of visualization, we project the estimated principle modes of K-means-PCA and MPPCA model from Euclidean space onto the sphere. 
Our model automatically separates
the sphere data into three groups, which aligns fairly well with the 
ground truth (Green: $200$; Blue: $289$; Black: $275$). For geodesics
in each cluster (ground truth in yellow and model estimate
in red), our results overlap better with the ground truth than others. 
This also indicates that our model can recover the parameters closer to
the truth (as shown in Tab.~\ref{tab:my_label}). In particular, the MBPGA model is able to automatically select an effective dimension of the principal subspaces to represent data variability.

\vspace{-8mm}
\begin{table*}[h]
    \footnotesize
    \caption{Comparison between ground truth parameters $\{\lambda_k, \pi_k, \tau_k\}$
     and the estimation of our model and baseline algorithms.}
    \label{tab:my_label}
    \centering
    \setlength{\tabcolsep}{+0.5mm}{
    \begin{tabular}{|c |c |c |c| }
    \hline
            & $\lambda_{k=1,2,3}$ & $\pi_{k=1,2,3}$ & $\tau_{k=1,2,3}$\\
         \hline
        Ground truth & (0.2, 0.01, 0)& (0.2618, 0.3783, 0.3599) & (277.7778, 123.4568, 69.4444) \\
         \hline
        K-means-PCA  & (0.1843, 0.0177, 0) & (0.2500, 0.3927, 0.3573) & NA  \\
        \hline
        MPPCA    & (0.5439, 0.0450, 0) & (0.2585, 0.3586, 0.3829) & (163.9344, 107.5269, 101.0101) \\
        \hline
        MPPGA   & (0.1901, 0.0099, 0) & (0.2618, 0.3783, 0.3599) & 
        (211.8783, 137.7593, 94.8111) \\
        \hline
        MBPGA   & (0.1905, 0, 0) & (0.2618, 0.3783, 0.3599) & (212.4965, 140.0511, 96.1169) \\
        \hline
    \end{tabular}}
\end{table*}

\begin{figure}
    \centering
    \includegraphics[width=\linewidth]{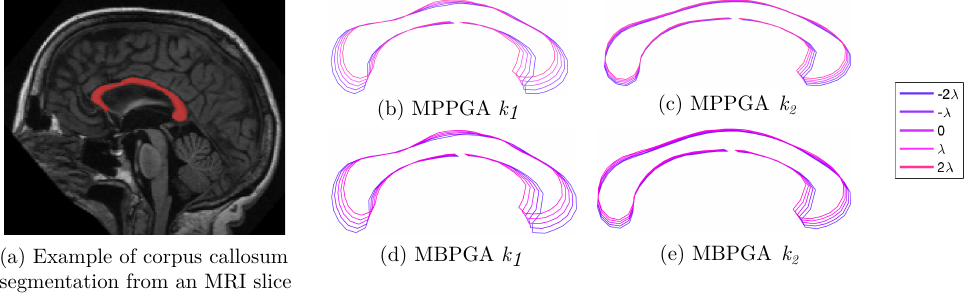}
    \caption{Corpus callosum shape variations (healthy $k_1$ vs. control $k_2$) along
the first principal geodesic ($- 2, -1, 0, 1, 2) \times \sqrt{\lambda}$
estimated by our model MPPGA and MBPGA.}
    \label{fig:6}
\end{figure}

Fig.~\ref{fig:6} demonstrates result of shape variations estimated 
by our model MPPGA and MBPGA. The corpus callosum shapes are 
automatically clustered into two different groups: healthy vs. control. 
An example of a segmented corpus callosum from brain MRI is shown in Fig.~\ref{fig:6}(a). Fig.~\ref{fig:6}(b) - Fig.~\ref{fig:6}(e) show shape variations generated from  points along the first principal geodesic: $\Exp(\mu,\alpha w^a)$, where $\alpha=- 2, -1, 0, 1, 2 \times \sqrt{\lambda})$, for $a = 1$. 
It is shown that the corpus callosum from healthy group is significantly 
larger than control group. Meanwhile, the anterior and posterior
ends of the corpus callosum show larger variation than the
mid-caudate, which is consistent with previous studies.

\begin{figure}[!h]
    \centering
    \includegraphics[width=.4\linewidth]{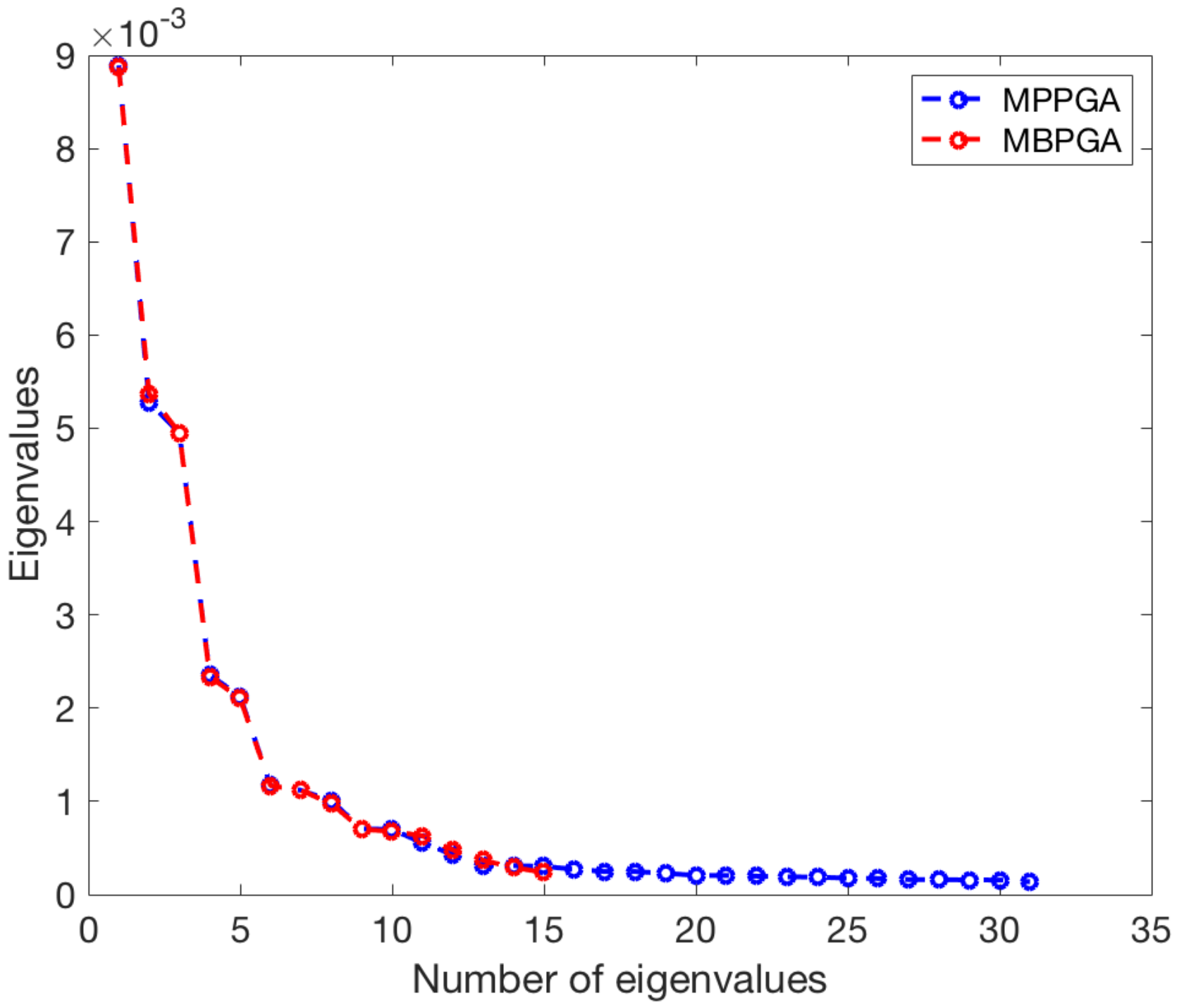}
    \caption{Eigenvalues estimated by MPPGA/ MBPGA on corpus callosum data.}
    \label{fig:eigenvalue}
\end{figure}

Fig.~\ref{fig:eigenvalue} shows fairly close eigenvalues estimated by MPPGA and MBPGA on corpus callosum data. Since the ARD prior introduced in MBPGA automatically suppresses irrelevant principal geodesics to zero, we have $15$ selected out of $128$ in total. 

\begin{figure}
    \centering
    \includegraphics[width=\linewidth]{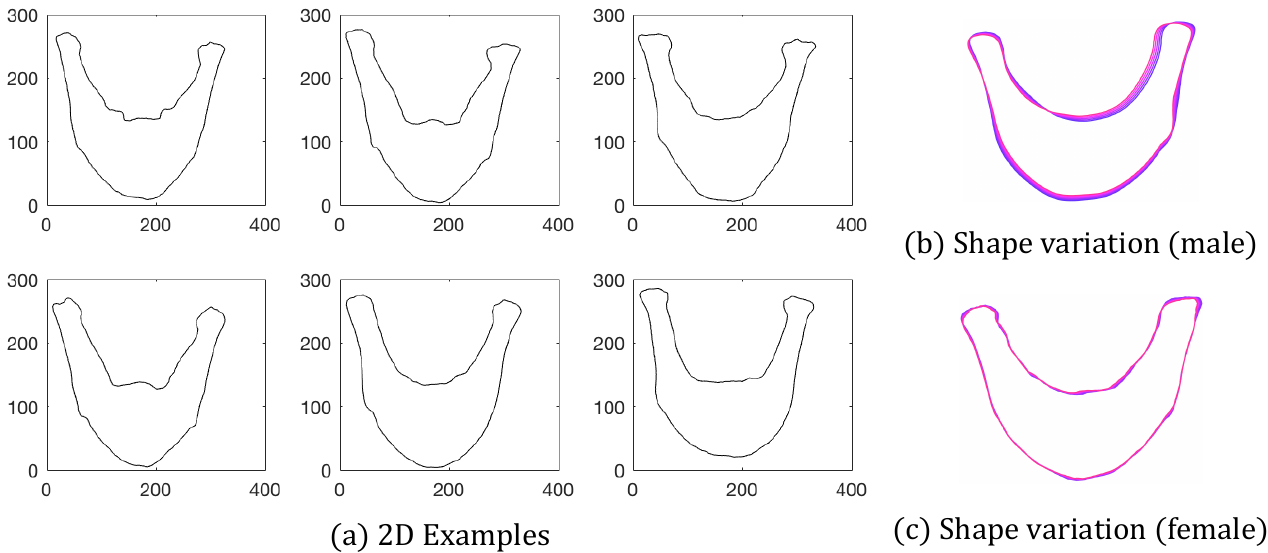}
    \caption{2D examples of mandible shape data and shape variations (male vs. female) along the first principal geodesic ($- 2, -1, 0, 1, 2) \times \sqrt{\lambda}$ estimated by MBPGA model.}
    \label{fig:mandible}
\end{figure}

We validate our MBPGA model to analyze the the mandible shape data ( visualization of 2D examples are shown in Fig.~\ref{fig:mandible}(a)) since MBPGA produces fairly close results as MPPGA, but with the functionality of automatic data dimensionality reduction. The MBPGA model reduces the original data dimension from $d=800$ to $d=70$. Fig.~\ref{fig:mandible}(b)(c) displays shape variations of mandibles from both male and female group. It clearly shows that generally male mandibles have larger variations than female mandibles, which is consistent with previous studies~\cite{chung2015unified}. In particular, male mandibles have a larger variation in the temporal crest and the base of mandible.


\section{Conclusion $\&$ Future Work}
We presented a mixture model of PGA (MPPGA) on general Riemannian manifolds. We developed an Expectation Maximization for maximum likelihood estimation of parameters including the underlying principal subspaces and automatic data clustering results. This work takes the first step to generalize mixture models of principal mode analysis to Riemannian manifolds. A Bayesian variant of MPPGA (MBPGA) was also discussed in this paper for automatic dimensionality reduction. This model is particularly useful, as it avoids singularities that are associated with maximum likelihood estimations by suppressing the irrelevant information, e.g., outliers or noises. Our proposed model also paves a way for new tasks on manifolds such as hierarchical clustering and classification. Notice that all experiments conducted in this paper are with the number of clusters $k$ being determined (e.g., healthy vs. control in corpus callosum data, or male vs. female in mandible data). For datasets with completely unknown clusters, current methods such as Elbow~\cite{ketchen1996application}, Silhouhette~\cite{kaufman1990partitioning}, and Gap statistic methods~\cite{tibshirani2001estimating} can be performed to determine the optimal number of clusters. This will be further investigated in our future work.
%
%
\bibliography{Sections/ref.bib}
\bibliographystyle{splncs04}
\end{document}